\documentclass[10pt,twocolumn,letterpaper]{article}

\usepackage{wacv}              

\usepackage{graphicx}
\usepackage{amsmath}
\usepackage{amssymb}
\usepackage{booktabs}
\usepackage{amsthm}
\usepackage{multirow}
\usepackage[accsupp]{axessibility}

\usepackage{pifont}

\usepackage{tikz}
\newtheorem{theorem}{Theorem}
\newtheorem{prop}[theorem]{Proposition}

\usepackage[pagebackref,breaklinks,colorlinks]{hyperref}

\usepackage[capitalize]{cleveref}
\crefname{section}{Sec.}{Secs.}
\Crefname{section}{Section}{Sections}
\Crefname{table}{Table}{Tables}
\crefname{table}{Tab.}{Tabs.}

\def\wacvPaperID{786} 
\def\confName{WACV}
\def\confYear{2024}

\begin{document}
\title{Hyperbolic vs Euclidean Embeddings in Few-Shot Learning: \\ Two Sides of the Same Coin}

\author{Gabriel Moreira\\
Language Technologies Institute\\
Carnegie Mellon University\\
{\tt\small gmoreira@cs.cmu.edu}
\and
Manuel Marques\\
Institute for Systems and Robotics\\
Instituto Superior Técnico\\
{\tt\small manuel@isr.tecnico.ulisboa.pt}
\and
João Paulo Costeira\\
Institute for Systems and Robotics\\
Instituto Superior Técnico\\
{\tt\small jpc@isr.tecnico.ulisboa.pt}
\and
Alexander Hauptmann\\
Language Technologies Institute\\
Carnegie Mellon University\\
{\tt\small alex@cs.cmu.edu}
}
\maketitle

\begin{abstract}
Recent research in representation learning has shown that hierarchical data lends itself to low-dimensional and highly informative representations in hyperbolic space. However, even if hyperbolic embeddings have gathered attention in image recognition, their optimization is prone to numerical hurdles. Further, it remains unclear which applications stand to benefit the most from the implicit bias imposed by hyperbolicity, when compared to traditional Euclidean features. In this paper, we focus on prototypical hyperbolic neural networks. In particular, the tendency of hyperbolic embeddings to converge to the boundary of the Poincaré ball in high dimensions and the effect this has on few-shot classification. We show that the best few-shot results are attained for hyperbolic embeddings at a common hyperbolic radius. In contrast to prior benchmark results, we demonstrate that better performance can be achieved by a fixed-radius encoder equipped with the Euclidean metric, regardless of the embedding dimension. 
\end{abstract}

\section{Introduction}
\label{sec:intro}
Hyperbolic embeddings first appeared in machine learning as an inductive bias for representing data assumed to have a \textit{latent hierarchical structure} \cite{Nickel2017PoincareRepresentations, Chamberlain2017NeuralSpace, Chami2019HyperbolicNetworks}, outperforming Euclidean-based approaches in community detection, link prediction and word embedding benchmarks \cite{Chen2021FullyNetworks}. Not long after, they were enthusiastically adopted by the computer vision community, finding applications in classification \cite{Guo2022ClippedClassifiers}, zero-shot/few-shot learning  \cite{Liu2020HyperbolicRecognition,Khrulkov2019HyperbolicEmbeddings}, contrastive learning \cite{Yue2023HyperbolicLearning, Ge2022HyperbolicObjects,Lin2022ContrastiveClustering}, segmentation \cite{GhadimiAtigh2022HyperbolicSegmentation} and VAEs \cite{Nagano2019ALearning,Mathieu2019ContinuousAuto-encoders,Skopek2020Mixed-curvatureAutoencoders}. The theoretical motivation underlying their use stems from the fact that hyperbolic geometry can theoretically capture complex hierarchical features that are otherwise  elusive in low-dimensional Euclidean space \cite{Nickel2017PoincareRepresentations,Nickel2018LearningGeometry,Chamberlain2017NeuralSpace}.

\begin{figure}
\centering
\input{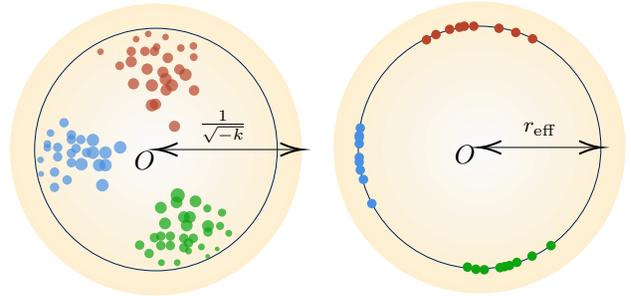}
\caption{Hyperbolic image embeddings in the Poincaré ball: expectation (left) versus reality (right). In high-dimensional hyperbolic space, the volume of a ball is concentrated near its surface where the hyperbolic metric varies monotonically with the angle. Thus, the hierarchy-revealing property of hyperbolic space is lost.}
\label{fig:example}
\end{figure}

In spite of the promising empirical results reported in the computer vision literature thus far, ascertaining which tasks can benefit the most from hyperbolic geometry is not yet clear. Up until recently, even classical cross-entropy based image classification in the Poincaré ball model of hyperbolic space was hardly on par with its Euclidean counterpart, much less competing with successful non-computer vision applications. Recent research \cite{Guo2022ClippedClassifiers} has linked this performance gap to embeddings converging towards the boundary of the ball. There, the Riemannian gradient vanishes, which effectively brings the optimization to a halt. It is known that the hyperboloid model outperforms the Poincaré ball partly because it circumvents many such numerical issues when learning low-dimensional word embeddings \cite{Nickel2018LearningGeometry}. In image recognition applications, where representations are high-dimensional, a possible fix is to clip the embeddings' magnitude \cite{Guo2022ClippedClassifiers}, which prevents the gradient underflow. Despite producing better classifiers, this technique does not prevent the embeddings from saturating all at the boundary of the smaller `clipped' ball. Such a structural fact was obscured by  performance indicators.

Our insight is that hyperbolic prototypical networks with a high-dimensional output space, typically in the order of $10^3$ in few-shot learning, tend to produce embeddings with the largest possible magnitude, as illustrated in Fig. \ref{fig:example}. In other words, instead of populating the entire space of the ball as expected, we observe that top performers place the visual embeddings near the border. In fact, the prototypical loss \cite{Snell2017PrototypicalLearning} is unbounded below.

We show that the hyperbolic measure of a high-dimensional ball is concentrated at its boundary, similarly to what happens in Euclidean space. This is corroborated by our experiments in popular few-shot datasets, which demonstrate that, regardless of magnitude clipping, in high-dimensional hyperbolic space, image embeddings lie at the maximum radius allowed. Thus, even if we employ a hyperbolic metric, the representations are restricted to a sphere where the distance between points varies monotonically with the angle between them. Such a scenario can be mirrored in Euclidean space since, in this case, there is no evidence of the property that sets hyperbolic space apart: its ability to encode hierarchical information in low dimensions.

In summary, we make a threefold contribution: 1) We show that, in high-dimensional hyperbolic space, the volume of a ball is concentrated at its boundary. This phenomenon, well established in Euclidean space, is often linked to the \textit{curse of dimensionality}; 2) We demonstrate empirically that the boundary saturation is actually the configuration wherein the current best hyperbolic few-shot results are attained; 3) Since the boundary of the space where embeddings lie is a hyperbolic sphere, we show that a fixed-radius Euclidean encoder is able to match and even improve the hyperbolic classification accuracy without any of the intricacies of Riemannian optimization. 

Prior benchmarks did not perform such a comparison, leading to the belief that hyperbolic embeddings were better suited towards few-shot recognition than traditional Euclidean features, by a considerable margin. Our findings present a stark contrast with previous Euclidean versus hyperbolic benchmarks \cite{Guo2022ClippedClassifiers,Khrulkov2019HyperbolicEmbeddings}, where a considerable difference between the two geometries was observed. 

Intuitively, our results suggest that individual images are simply instances of visual semantic relations and do not convey higher level concepts represented by a semantic hierarchy. In fact, while thinking of a concept, say a bird, we have great difficulty in finding one single image that constitutes a `parent' of the whole class. In this view, the best classifier should represent all images as leaves of a tree, in other words, at the same radius in the Poincaré ball.

\section{Related work}
\label{sec:related_work}
The use of hyperbolic space in machine learning can be traced back to earlier works \cite{Krioukov2010HyperbolicNetworks,Boguna2009NavigabilityNetworks} that set forth the hypothesis that complex networks may be thought of as discrete observations of a latent smooth metric space. For data points which are semantically meaningful and thus, amenable to being hierarchically categorized, hyperbolic geometry arises naturally as a consequence of node similarity distances. This duality between hierarchies, or more generally trees, and hyperbolic geometry was also shown by Sarkar \cite{Sarkar2011LowPlane} who proposed a combinatorial algorithm for embedding any tree in the Poincaré ball, while preserving the finite graph metric. 

\paragraph{Symbolic data} As a means to circumvent the limitations of Euclidean space in capturing complex symbolic patterns, Chamberlain \etal \cite{Chamberlain2017NeuralSpace} described a hyperbolic version of the celebrated word2vec architecture \cite{Mikolov2013DistributedCompositionality} in the Poincaré ball model. The authors showed that community detection and node attribute prediction in low-dimensional hyperbolic space outperforms methods based on higher-dimensional Euclidean embeddings. Nickel and Kiela \cite{Nickel2017PoincareRepresentations}, adopting a similar model, reported similar findings for WordNet reconstruction and link prediction tasks. Unlike the former approach however, the authors derived a stochastic Riemannian optimization method. Subsequent work \cite{Nickel2018LearningGeometry}, motivated by the numerical hurdles presented by optimization in the Poincaré ball, proposed supplanting it for the Lorentz model, also known as the hyperboloid model. It was through the use of this model that Sala \etal \cite{Sala2018RepresentationEmbeddings} derived a matrix factorization-based hyperbolic multidimensional scaling algorithm (MDS) and Chami \etal designed hyperbolic graph convolutions \cite{Chami2019HyperbolicNetworks}.

\paragraph{Image recognition}  In image recognition, hyperbolic representation spaces have been employed with different end goals. One of them being the case wherein a hierarchy is established before-hand and the desiderata is for the representation space to be structured accordingly. Such was the basis for the zero-shot architecture proposed by Liu \etal \cite{Liu2020HyperbolicRecognition}, where WordNet was used as prior hierarchical knowledge of ImageNet, and its embeddings interpreted as class prototypes. Conversely, actual hyperbolic prototypical learning approaches have been set forth, such as ideal hyperbolic prototypes, proposed by Atigh \etal \cite{Atigh2021HyperbolicPrototypes}, making use of Busemann functions. More closely related to the original prototypical learning paper \cite{Snell2017PrototypicalLearning} is the work by Khrulkov \etal \cite{Khrulkov2019HyperbolicEmbeddings}, where the authors retain the entire architecture, replacing simply the Euclidean metric by the Poincaré one, which yielded superior results.  Advances have also been made in classical image classification. Based on empirical evidence suggesting that hyperbolic networks were not on par with their Euclidean counterparts, Guo \etal \cite{Guo2022ClippedClassifiers} attributed this to gradient underflow arising from the inverse Riemannian metric tensor and proposed clipping the magnitude of the Euclidean features before these are projected to the hyperbolic manifold. Finally, hyperbolic space has also been applied to image segmentation \cite{GhadimiAtigh2022HyperbolicSegmentation}, VAEs \cite{Mathieu2019ContinuousAuto-encoders,Nagano2019ALearning,Skopek2020Mixed-curvatureAutoencoders} and contrastive learning \cite{Yue2023HyperbolicLearning, Ge2022HyperbolicObjects,Lin2022ContrastiveClustering}.

The remainder of this paper is organized as follows. In Section \ref{sec:hyperbolic_space} we lay out a brief summary of hyperbolic geometry, namely the hyperboloid and Poincaré ball models. In Section \ref{sec:theory} we put forward the reasoning behind why we should expect hyperbolic image encoders to behave similarly to fixed-radius Euclidean ones. Finally, we present empirical results to validate our approaches in Section \ref{sec:experiments} and draw concluding remarks subsequently.

\section{Hyperbolic space}
\label{sec:hyperbolic_space}

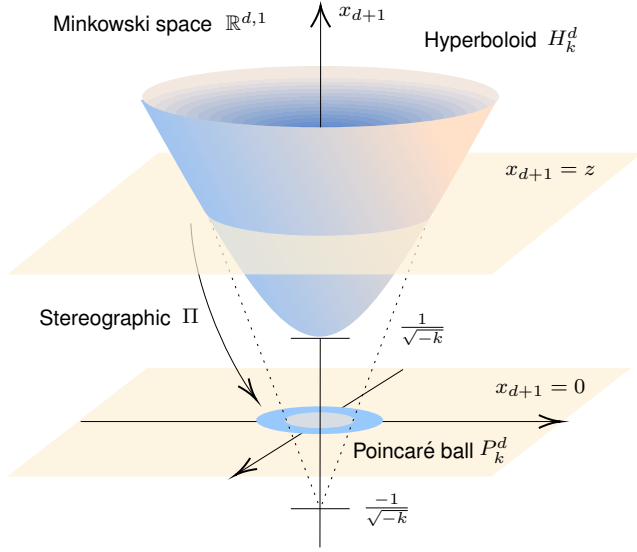
\begin{figure}
\centering
  
\tikzset {_hspm0mmty/.code = {\pgfsetadditionalshadetransform{ \pgftransformshift{\pgfpoint{13.5 bp } { -37.5 bp }  }  \pgftransformrotate{-336 }  \pgftransformscale{2 }  }}}
\pgfdeclarehorizontalshading{_u6ja8bmjh}{150bp}{rgb(0bp)=(0.47,0.72,1);
rgb(37.5bp)=(0.47,0.72,1);
rgb(62.5bp)=(1,0.88,0.79);
rgb(100bp)=(1,0.88,0.79)}

  
\tikzset {_km2gm8rfr/.code = {\pgfsetadditionalshadetransform{ \pgftransformshift{\pgfpoint{0 bp } { 328 bp }  }  \pgftransformscale{1 }  }}}
\pgfdeclareradialshading{_rjm0blf32}{\pgfpoint{0bp}{-360bp}}{rgb(0bp)=(0.31,0.5,0.79);
rgb(0.44224330357142855bp)=(0.31,0.5,0.79);
rgb(25bp)=(0.95,0.91,0.88);
rgb(400bp)=(0.95,0.91,0.88)}
\tikzset{every picture/.style={line width=0.3pt}} 

\begin{tikzpicture}[x=0.75pt,y=0.75pt,yscale=-1,xscale=1]

\draw  [draw opacity=0][fill={rgb, 255:red, 253; green, 240; blue, 213 }  ,fill opacity=0.7 ] (227.2,82.33) -- (471.38,82.33) -- (433.3,113.03) -- (189.11,113.03) -- cycle ;
\draw  [draw opacity=0][fill={rgb, 255:red, 253; green, 240; blue, 213 }  ,fill opacity=0.7 ] (221.36,191) -- (465.56,191) -- (395.23,245.73) -- (151.03,245.73) -- cycle ;
\draw    (350.59,191.73) -- (267.69,243.93) ;
\draw [shift={(266,245)}, rotate = 327.8] [color={rgb, 255:red, 0; green, 0; blue, 0 }  ][line width=0.75]    (10.93,-3.29) .. controls (6.95,-1.4) and (3.31,-0.3) .. (0,0) .. controls (3.31,0.3) and (6.95,1.4) .. (10.93,3.29)   ;
\draw    (428,218) -- (188,218) ;
\draw [shift={(430,218)}, rotate = 180] [color={rgb, 255:red, 0; green, 0; blue, 0 }  ][line width=0.75]    (10.93,-3.29) .. controls (6.95,-1.4) and (3.31,-0.3) .. (0,0) .. controls (3.31,0.3) and (6.95,1.4) .. (10.93,3.29)   ;
\draw    (323.67,176) -- (294,176) ;
\draw  [draw opacity=0][fill={rgb, 255:red, 152; green, 201; blue, 253 }  ,fill opacity=1 ] (276.5,217.33) .. controls (276.5,213.47) and (290.83,210.33) .. (308.5,210.33) .. controls (326.17,210.33) and (340.5,213.47) .. (340.5,217.33) .. controls (340.5,221.2) and (326.17,224.33) .. (308.5,224.33) .. controls (290.83,224.33) and (276.5,221.2) .. (276.5,217.33) -- cycle ;
\draw    (323.67,262) -- (294,262) ;
\draw  [draw opacity=0][fill={rgb, 255:red, 224; green, 224; blue, 224 }  ,fill opacity=0.88 ] (291,217.33) .. controls (291,215.12) and (298.84,213.33) .. (308.5,213.33) .. controls (318.16,213.33) and (326,215.12) .. (326,217.33) .. controls (326,219.54) and (318.16,221.33) .. (308.5,221.33) .. controls (298.84,221.33) and (291,219.54) .. (291,217.33) -- cycle ;
\draw  [dash pattern={on 0.84pt off 2.51pt}]  (364.59,112.07) -- (308.83,262) ;
\draw  [dash pattern={on 0.84pt off 2.51pt}]  (251.59,112.07) -- (308.83,262) ;
\draw    (243.59,118) .. controls (244.57,145.44) and (258.99,182.48) .. (275.57,205.6) ;
\draw [shift={(276.59,207)}, rotate = 233.53] [color={rgb, 255:red, 0; green, 0; blue, 0 }  ][line width=0.75]    (10.93,-3.29) .. controls (6.95,-1.4) and (3.31,-0.3) .. (0,0) .. controls (3.31,0.3) and (6.95,1.4) .. (10.93,3.29)   ;
\draw [draw opacity=0][shading=_u6ja8bmjh,_hspm0mmty]   (218.21,55.57) .. controls (308.21,215.57) and (309,215) .. (399,54) ;
\draw  [draw opacity=0][shading=_rjm0blf32,_km2gm8rfr] (219,54) .. controls (219,45.46) and (259.29,38.53) .. (309,38.53) .. controls (358.71,38.53) and (399,45.46) .. (399,54) .. controls (399,62.54) and (358.71,69.47) .. (309,69.47) .. controls (259.29,69.47) and (219,62.54) .. (219,54) -- cycle ;
\draw    (309,10.03) -- (309,69.47) ;
\draw [shift={(309,8.03)}, rotate = 90] [color={rgb, 255:red, 0; green, 0; blue, 0 }  ][line width=0.75]    (10.93,-4.9) .. controls (6.95,-2.3) and (3.31,-0.67) .. (0,0) .. controls (3.31,0.67) and (6.95,2.3) .. (10.93,4.9)   ;
\draw  [draw opacity=0][fill={rgb, 255:red, 253; green, 240; blue, 213 }  ,fill opacity=0.7 ] (433.3,113.03) -- (395.21,143.73) -- (151.03,143.73) -- (189.11,113.03) -- (251.52,113.03) .. controls (254.52,119.33) and (279.07,124.23) .. (308.92,124.23) .. controls (338.76,124.23) and (363.31,119.33) .. (366.31,113.03) -- (433.3,113.03) -- cycle ;
\draw    (308.83,176) -- (308.67,281.53) ;

\draw (317,7.4) node [anchor=north west][inner sep=0.75pt]  [font=\footnotesize]  {$x_{d+1}$};
\draw (422,17.4) node [anchor=north west][inner sep=0.75pt]  [font=\footnotesize]  {$H_{k}^{d}$};
\draw (401,88) node [anchor=north west][inner sep=0.75pt]  [font=\footnotesize]  {$x_{d+1} =z$};
\draw (260,10) node [anchor=north west][inner sep=0.75pt]  [font=\footnotesize]  {$\mathbb{R}^{d,1}$};
\draw (347,162.4) node [anchor=north west][inner sep=0.75pt]  [font=\footnotesize]  {$\frac{1}{\sqrt{-k}}$};
\draw (329,253.4) node [anchor=north west][inner sep=0.75pt]  [font=\footnotesize]  {$\frac{-1}{\sqrt{-k}}$};
\draw (388,224.52) node [anchor=north west][inner sep=0.75pt]  [font=\footnotesize]  {$P_{k}^{d}$};
\draw (396,195.4) node [anchor=north west][inner sep=0.75pt]  [font=\footnotesize]  {$x_{d+1} =0$};
\draw (238,160) node [anchor=north west][inner sep=0.75pt]  [font=\footnotesize]  {$\Pi $};
\draw (360,20) node [anchor=north west][inner sep=0.75pt]  [font=\footnotesize\sffamily] [align=left] {{Hyperboloid}};
\draw (166,160) node [anchor=north west][inner sep=0.75pt]  [font=\footnotesize\sffamily] [align=left] {{Stereographic}};
\draw (324,227) node [anchor=north west][inner sep=0.75pt]  [font=\footnotesize\sffamily] [align=left] {{Poincaré ball}};
\draw (173,12) node [anchor=north west][inner sep=0.75pt]  [font=\footnotesize\sffamily] [align=left] {{Minkowski space}};

\end{tikzpicture}
\vspace{-0.3cm}
\caption{Minkowski ambient space $\mathbb{R}^{d,1}$, hyperboloid $H_k^d$, stereographic projection $\Pi$ and Poincaré ball model $P_k^d$.}
\label{fig:hmodels}
\end{figure}

Much like Euclidean space, hyperbolic space is isotropic and completely defined by its constant curvature, which unlike the former is negative everywhere. Several isomorphic models of hyperbolic space exist and in the machine learning literature, the most commonly adopted are the hyperboloid, also known as the Lorentz model \cite{Nickel2018LearningGeometry,Chami2019HyperbolicNetworks,Chen2021FullyNetworks}, and the Poincaré ball \cite{Chamberlain2017NeuralSpace,Khrulkov2019HyperbolicEmbeddings, Nickel2017PoincareRepresentations,Guo2022ClippedClassifiers,Atigh2021HyperbolicPrototypes}, obtained from a stereographic projection of the former.

Throughout this paper we will be working with the Poincaré ball model of hyperbolic space, which can be derived from the hyperboloid model as follows. Consider the Minkowski space $\mathbb{R}^{d,1}:=\{\mathbf{x} = (x_1,\dots,x_{d+1}) \in \mathbb{R}^{d}\times\mathbb{R}\}$, together with the bilinear form (Lorentz pseudometric)
\begin{equation}
    \langle \mathbf{x}, \mathbf{y} \rangle_L := \sum_{i=1}^d x_i y_i - x_{d+1} y_{d+1}.
    \label{eq:lorentz_pseudometric}
\end{equation}
The form $\langle \mathbf{x},\mathbf{x} \rangle_L$ is known as a Lorentz quadratic form. While not positive-definite everywhere in $\mathbb{R}^{d,1}$ and thus, not a metric, it is positive-definite when restricted to the upper sheet of the $d$-hyperboloid
\begin{equation}
    H^d_k := \left\{\mathbf{x} \in \mathbb{R}^{d,1} \;|\; \langle \mathbf{x},\mathbf{x} \rangle_L = \frac{1}{k},\; x_{d+1} > 0\right\},
    \label{eq:definition_hyperboloid}
\end{equation}
with curvature $k < 0$. We can write $H^d_k$ in local coordinates $u_1,\dots, u_d$ via the inclusion map $\phi: H^d_k \to \mathbb{R}^{d,1}$
\begin{equation}
    \phi(\mathbf{u}) := \left(u_1,\dots,u_d,\sqrt{\|\mathbf{u}\|_2^2 - \frac{1}{k}}\right).
    \label{eq:inclusion_map}
\end{equation}

\paragraph{Poincaré model} The Poincaré ball model with sectional curvature $k<0$, denoted as $P^d_k$, is obtained via the stereographic projection $\Pi : H^d_k \to P^d_k$ of the $d$-hyperboloid through the origin (see Figure \ref{fig:hmodels})
\begin{equation}
    \Pi(\mathbf{x}) := \left(\frac{x_1}{1+\sqrt{-k}x_{d+1}},\dots,\frac{x_d}{1+\sqrt{-k}x_{d+1}}\right).
    \label{eq:stereographic_projection}
\end{equation}
This yields the ball 
\begin{equation}
    B^d_k=\Pi(H_k^d)=\left\{x\in\mathbb{R}^d : \|\mathbf{x}\|_2^2 < -\frac{1}{k}\right\},
\end{equation}
where the boundary corresponds to infinity. The metric tensor in the canonical basis of $\mathbb{R}^{d}$ is  $\lambda(\mathbf{u})^2\mathbf{I}_{d}$, with $\lambda(\mathbf{u})=2/(1+k\|\mathbf{u}\|_2^2)$. The inverse projection $\Pi^{-1} : P^d_k \to H^d_k$ takes the form
\begin{equation}
    \Pi^{-1}(\mathbf{u})=\left(\lambda(\mathbf{u})\mathbf{u},\frac{1}{\sqrt{-k}}(\lambda(\mathbf{u})-1)\right).
\end{equation}
Since $\lambda(\mathbf{u})\to \infty$ as $\|\mathbf{u}\|_2\to 1/\sqrt{-k}$, for numerical reasons the effective radius of the ball is usually capped \cite{Ganea2018HyperbolicNetworks,Khrulkov2019HyperbolicEmbeddings} \ie for $\epsilon > 0$,
\begin{equation}
    r_{\mathrm{eff}} := \frac{1-\epsilon}{\sqrt{-k}}.
    \label{eq:radius_eff}
\end{equation}

The Poincaré exponential map at the origin $\mathrm{Exp}_0(\mathbf{v}): T P_k^d \to P_k^d$ allows us to project a tangent vector back to the ball. This map is the element responsible for converting an Euclidean neural network to a hyperbolic one and reads
\begin{align}
    \mathrm{Exp}_0(\mathbf{v}) = \tanh\left(\sqrt{-k}\|\mathbf{v}\|_2\right)\frac{\mathbf{v}}{\sqrt{-k}\|\mathbf{v}\|_2}.
    \label{eq:poincare_exp_map}
\end{align}
Finally, the Poincaré geodesic (shortest-path) distance between any $\mathbf{x}$ and $\mathbf{y}\in P_k^d$ is given by
\begin{equation}
    d_{P_k^d}(\mathbf{x},\mathbf{y}) := \frac{2}{\sqrt{-k}}\mathrm{atanh}\left(\sqrt{-k}\|-\mathbf{x}\oplus \mathbf{y}\|_2\right),
    \label{eq:poincare_distance}
\end{equation}
where $\oplus$ denotes the Möbius addition,
\begin{equation}
    \mathbf{x}\oplus\mathbf{y} = \frac{(1-2k\langle \mathbf{x},\mathbf{y}\rangle - k\|\mathbf{y}\|_2^2)\mathbf{x} + (1+k\|\mathbf{x}\|_2^2)\mathbf{y}}{1-2k\langle \mathbf{x},\mathbf{y} \rangle + k^2\|\mathbf{x}\|_2^2\|\mathbf{y}\|_2^2}.
\end{equation}

\paragraph{Poincaré image encoder} Given an Euclidean backbone $f$ with parameters $\theta$, we consider hyperbolic image encoders of the type
\begin{equation}
    h(\mathbf{x};\theta) = \mathrm{Exp}_0^P \left(f(\mathbf{x}; \theta)\right).
\end{equation}
The clipping strategy put forward in \cite{Guo2022ClippedClassifiers} consists of setting a maximum magnitude $c$ for $f(\mathbf{x}; \theta)$, restricting the Poincaré ball to a ball with effective radius $r_\mathrm{eff}=\mathrm{tanh}(\sqrt{-k}c)/\sqrt{-k}$. During training, for a suitable loss function $L$, and letting $\mathbf{z} := h(\mathbf{x};\theta) \in P_{k}^d$, the Euclidean gradient $\nabla_{\mathbf{z}} L$ may be backpropagated as is (see implementation by \cite{Khrulkov2019HyperbolicEmbeddings}), or it may be converted to a Riemannian gradient via scaling by the inverse metric tensor $\mathrm{grad}_{\mathbf{z}} L = \lambda(\mathbf{z})^{-2} \nabla_{\mathbf{z}} L(\mathbf{z})$, as in \cite{Guo2022ClippedClassifiers}. 

\section{Few-shot in the boundary}
\label{sec:theory}
It is known that in hyperbolic neural networks, embeddings are prone to converge to the boundary of $P_k^d$ (in practice, the effective radius $r_\mathrm{eff}$). We focus on the fact that, from a measure concentration argument, at the dimensions typically employed in few-shot learning, hyperbolic embeddings should concentrate at the boundary, regardless of whether their magnitude is clipped. 

\paragraph{Hyperbolic prototypical learning} We assume an image dataset $\mathcal{I}$ with $C$ semantic classes. If $\mathcal{M}$ is a manifold and $f_\theta : \mathcal{I} \to \mathcal{M}$ an image encoder parameterized by $\theta$, a typical prototypical learning approach \cite{Snell2017PrototypicalLearning} models the probability of $\mathbf{z}_i$ being of class $c$ as
\begin{equation}
    p(c|\mathbf{z}_i) = \frac{\exp(-d_{\mathcal{M}}(\mathbf{w}_{c}, f(\mathbf{z}_i;\theta)))}{\sum_{k}{\mathrm{exp}(-d_{\mathcal{M}}(\mathbf{w}_{k}, f(\mathbf{z}_i;\theta))}},
\end{equation}
where $\mathbf{w}_{c}\in\mathcal{M}$ is the centroid of the $c$-th class. In classical prototypical networks, the squared Euclidean norm $\ell_2^2$ is chosen as $d_\mathcal{M}$ \cite{Snell2017PrototypicalLearning}. In Poincaré networks, the geodesic distance (\ref{eq:poincare_distance}) is used instead. For simplicity, let $\mathbf{x}_i=f(\mathbf{z}_i;\theta)$ and assume we are optimizing over $\mathbf{x}$ instead of $\theta$. The negative log-likelihood associated with the $i$-th image is then
\begin{align}
    L_i = d_\mathcal{M}\left(\mathbf{w}_{c},\mathbf{x}_i\right)
    + \log\sum_{k} e^{-d_\mathcal{M}(\mathbf{w}_{k},\mathbf{x}_i)}
    \label{eq:contrastive_loss}
\end{align}
and the loss is $L=\sum_i L_i$. We can easily verify that $L$ is unbounded below. Pick $C$ directions in $\mathbb{R}^d$, one for each class centroid $\mathbf{w}_c$. Then set the direction of each embedding to match that of its class $\mathbf{x}_i = r\mathbf{w}_{c}$ for a certain $r>0$. The first term of (\ref{eq:contrastive_loss}) is automatically zero and the second goes to $-\infty$ as the embeddings approach the boundary $r\to 1/\sqrt{-k}$. This justification could just as well be applied to Euclidean space, with the difference that we would require the output of the encoder to be unbounded. This is not necessary in the Poincaré model. In fact, the Poincaré exponential map at the origin (\ref{eq:poincare_exp_map}) used to map points from the tangent space $\mathbb{R}^d$ back to the manifold, is a scaling factor that allows us to approach the boundary at an exponential rate.

\paragraph{High-dimensional hyperbolic space} In $d$-dimensional Euclidean space, the ratio of the volume over the area of a ball decreases with $1/d$ \ie, in high-dimensional Euclidean space the volume is \textit{concentrated} close to the surface. In hyperbolic space, the same thing happens.

\begin{prop}[Hyperbolic measure concentration]
For large $d$, the volume of a hyperbolic ball is concentrated close to its boundary .
\begin{proof}
From III.4 \cite{chavel2006}, the volume of a hyperbolic ball of (hyperbolic) radius $r$ is
\begin{equation}
    V_k(r) = \frac{2\pi^{d/2}}{\Gamma(d/2)} \int_0^{r} \left(\frac{\sinh(\sqrt{-k}t)}{\sqrt{-k}}\right)^{d-1} dt
\end{equation}
and its area is given by
\begin{equation}
    A_k(r) = \frac{2\pi^{d/2}}{\Gamma(d/2)} \left(\frac{\sinh(\sqrt{-k}r)}{\sqrt{-k}}\right)^{d-1}.
\end{equation}
Hence,
\begin{align}
    \frac{V_k(r)}{A_k(r)} &= \left(\frac{\sinh(\sqrt{-k}r)}{\sqrt{-k}}\right)^{1-d} \int_0^{r} \left(\frac{\sinh(\sqrt{-k}t)}{\sqrt{-k}}\right)^{d-1} dt \nonumber \\
    &= \int_0^{r} \left(\frac{\sinh(\sqrt{-k}t)}{\sinh(\sqrt{-k}r)}\right)^{d-1} dt.
\end{align}
For any $d > 1$, the function $\frac{\sinh(\sqrt{-k}t)}{\sinh(\sqrt{-k}r)}$ is convex for $t > 0$, and thus, for $t \in [0,r]$, it is bounded above by $t/r$. From
\begin{equation}
    0 \leq \left(\frac{\sinh(\sqrt{-k}t)}{\sinh(\sqrt{-k}r)}\right)^{d-1} \leq \frac{t^{d-1}}{r^{d-1}},
\end{equation}
the integral can be bounded above as
\begin{equation}
    \frac{V_k(r)}{A_k(r)} \leq  \int_0^{r} \frac{t^{d-1}}{r^{d-1}} dt = \frac{r}{d}.
\end{equation}
Thus, $\frac{V_k(r)}{A_k(r)} \to 0$ as $d\to \infty$.
\end{proof}
\end{prop}

Note that this result holds for hyperbolic space in general, namely the Poincaré ball and the hyperboloid models commonly employed in machine learning. These arguments lead us to our hypothesis: given the high dimensionality of the Poincaré ball used in the hyperbolic few-shot literature, embeddings should lie at, or close to, $r_\mathrm{eff}$. If this holds, it brings into question the hyperbolicity of the learnt representation space. In fact, a hyperbolic $(d-1)$-sphere containing embeddings at $r_\mathrm{eff}$ from the origin is isometric to an Euclidean $(d-1)$-sphere of radius
\begin{equation}
     \frac{2/\sqrt{-k}+2r_\mathrm{eff}}{1/\sqrt{-k}-r_\mathrm{eff}}.
     \label{eq:euclidean_radius}
\end{equation}
This can be shown by attending to the fact that any $(d-1)$-sphere in hyperbolic space with hyperbolic radius $r$ is isometric to an Euclidean $(d-1)$-sphere with radius
$\frac{1}{\sqrt{-k}}\sinh(r\sqrt{-k})$. If embeddings lie at $r_\mathrm{eff}$ from the origin, their hyperbolic radius is
\begin{equation}
    \frac{1}{\sqrt{-k}}\log \left(\frac{1/\sqrt{-k}+r_\mathrm{eff}}{1/\sqrt{-k}-r_\mathrm{eff}}\right)
\end{equation}
and (\ref{eq:euclidean_radius}) follows. This isometry, in turn, implies that the representation space is not negatively curved.

\begin{figure*}
    \centering
    \includegraphics[width=0.31\linewidth,trim={0cm 0cm 1.8cm 0cm},clip]{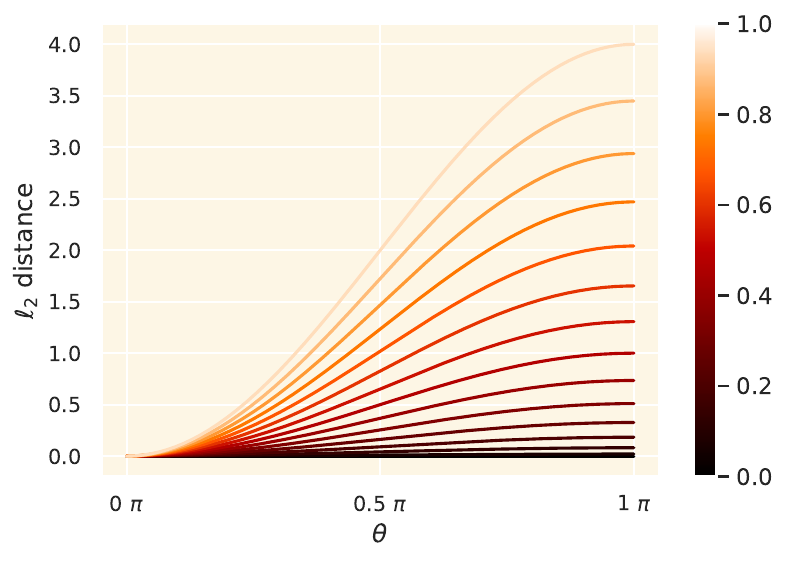}
    \includegraphics[width=0.31\linewidth,trim={0cm 0cm 1.8cm 0cm},clip]{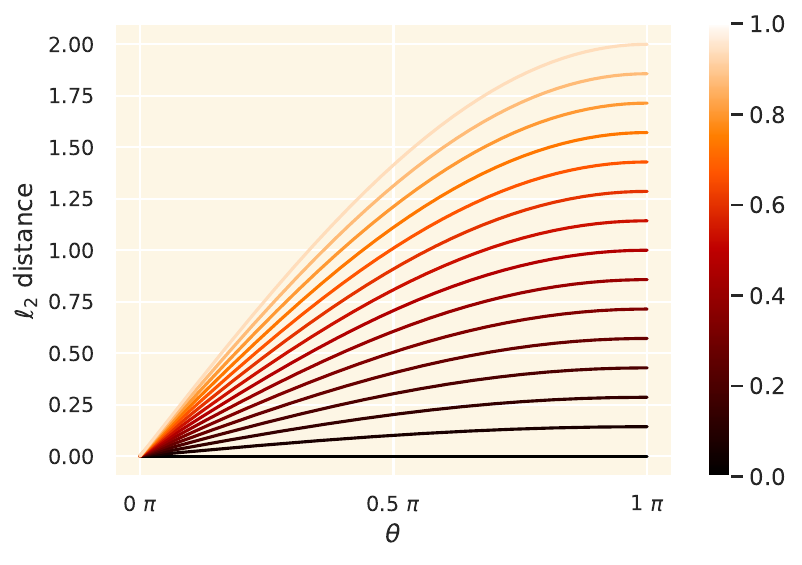}
    \includegraphics[width=0.35\linewidth,trim={0cm 0cm 0cm 0cm},clip]{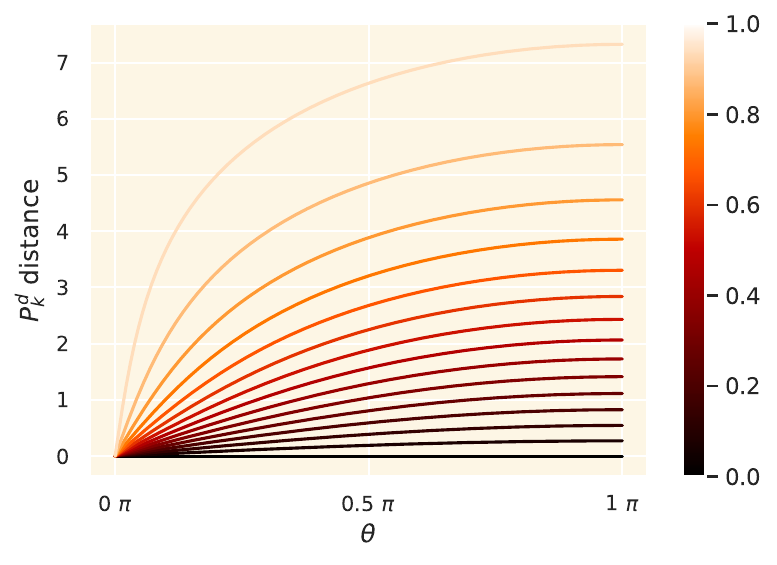}
    \caption{Comparison of metrics at different distances from the origin from 0 to $r_\mathrm{eff}=0.99$ (color encodes fraction of $r_\mathrm{eff}$). Left: squared Euclidean distance ($\ell_2^2$). Middle: Euclidean distance ($\ell_2$). Right: Poincaré distance ($d_{P_k^d}$) for $k=-1$.}
    \label{fig:distance_plots}
\end{figure*}

\paragraph{The Euclidean vs hyperbolic disparity} Assume our hypothesis holds and the embeddings all concentrate at $r_\mathrm{eff}$. While the metric (\ref{eq:poincare_distance}) employed in prior work \cite{Khrulkov2019HyperbolicEmbeddings,Gu2019LearningSpaces} is hyperbolic, it only encodes the angular distance. In fact, consider two embeddings $\mathbf{x},\mathbf{y}$ in $P_k^d$ such that $\|\mathbf{x}\|_2 = \| \mathbf{y}\|_2 = r_\mathrm{eff}$ and $\angle(\mathbf{x},\mathbf{y}) = \alpha$. The hyperbolic distance between them is
\begin{equation}
    d_{P_k^d}(\mathbf{x},\mathbf{y}) = \frac{1}{\sqrt{-k}}\mathrm{acosh}\left(a + b \cos(\alpha)\right),
    \label{eq:angular_poincare_dist}
\end{equation}
where $b = \frac{4kr_\mathrm{eff}^2}{(1+k r_\mathrm{eff}^2)^2}$ and $a=\left(\frac{1-k r_\mathrm{eff}^2}{1+k r_\mathrm{eff}^2}\right)^2$ (see supplemental material), which up to a multiplicative factor, depends only on $\theta$ and $k r_\mathrm{eff}^2$. A comparison between this metric, the $\ell_2$ and the $\ell_2^2$ is shown in Fig. \ref{fig:distance_plots}. As we observe on the rightmost plot, the derivative of $d_{P_k^d}$ at $\alpha\to 0$ surges as we approach the boundary. As this happens, the penalty applied to the angular distance $\alpha$ increases. Surprisingly, the Euclidean metric $\ell_2$ (center plot) is not that different from the hyperbolic. In fact, it becomes a good approximation far from the boundary. In contrast, consider the squared Euclidean distance $\ell_2^2$ depicted in the leftmost plot, often employed in prototypical networks since it defines a Bregman divergence \cite{Snell2017PrototypicalLearning}. Its derivative approaches $0$ as $\alpha\to 0$. 

Prior work conducted on hyperbolic prototypical learning benchmarked Euclidean space via $\ell_2^2$. Based on the previous paragraph, we propose instead fixed-radius Euclidean embeddings with the $\ell_2 \propto \sqrt{1-\cos(\alpha)}$ metric. Given a radius hyperparameter $r=1/\sqrt{k}$ ($k>0$  is a spherical curvature), and the Euclidean backbone $f(\cdot;\theta)$, the embeddings fed to the prototypical loss (\ref{eq:contrastive_loss}) are computed as
\begin{equation}
    r \frac{f(\mathbf{x};\theta)}{\|f(\mathbf{x};\theta) \|_2}.
    \label{eq:rescaling}
\end{equation}
As our experiments show, this  Euclidean architecture is ahead of the hyperbolic few-shot state-of-the-art in most of the experiments conducted.

\section{Experiments}
\label{sec:experiments}
We conducted our empirical evaluation in the CUB-200-2011 \cite{Wah2011TheDataset} and MiniImageNet \cite{Vinyals2016MatchingLearning} few-shot recognition tasks, in a prototypical learning regime. 

\paragraph{Architecture} We used a 4-layer ConvNet as backbone, with variable output dimension equal to $d$. These embeddings were then projected to the Poincaré ball through the exponential map at the origin (\ref{eq:poincare_exp_map}) and through magnitude rescaling in the Euclidean case (\ref{eq:rescaling}). In the latter, the radius is a hyperparameter, similarly to the curvature in the former. In the 5-shot 5-way setting, Poincaré centroids are computed through the Einstein midpoint closed-form expression set forth in \cite{Khrulkov2019HyperbolicEmbeddings}. In the Euclidean and the fixed-radius case, the Euclidean centroid was used. In terms of preprocessing, instead of cropping a centered square patch and resizing it to $84\times 84$ during inference as done in \cite{Khrulkov2019HyperbolicEmbeddings}, we padded the cropped bounding boxes in the case of CUB, and the original images in MiniImageNet, so as to obtain a square image and then resized them. All other data augmentations were the same and can be consulted in the supplemental material.

\paragraph{Training} All encoders were trained for 200 episodes, where each episode consisted of 15 queries, each of which associated with $s$-shot and $w$-way images. Adam was used with an initial learning rate of $10^{-3}$ on CUB and $5\times 10^{-3}$ on MiniImageNet. Similarly to prior work, a step LR scheduler was used (see supplemental material). Both on the CUB dataset and on MiniImageNet, we used the train-validation-test split from \cite{Khrulkov2019HyperbolicEmbeddings}. In the former, 100 classes were used for training, 50 for validation and 50 for testing, and the shot-way was the same for training and inference \ie, 1-shot 5-way and 5-shot 5-way. In the latter, the 100 class split was 64-16-20. The 1-shot 5-way model was trained in a 1-shot 30-way regime, and the 5-shot 5-way model in a 1-shot 20-way setting. 

\paragraph{Results} The CUB\_200\_2011 results are shown in Tables \ref{table:results_cub_1s5w} and  \ref{table:results_cub_5s5w}. Those of MiniImageNet are presented in Tables \ref{table:results_mini_1s5w} and \ref{table:results_mini_5s5w}. Given that hyperbolic space is often praised for its ability to encode information in few dimensions, for each dataset we present the test accuracy computed for four embedding dimensions $d=$ 128, 256, 512 and 1024. For hyperbolic and fixed-radius architectures the minimum, average and maximum $\ell_2$ embedding norms are reported as $r_\mathrm{min}$, $r_\mathrm{avg}$ and $r_\mathrm{max}$, respectively. The traditional Euclidean architecture used in the literature as the Euclidean baseline (ProtoNet) is denoted as $(\mathbb{R}^d,\ell_2^2)$, the Poincaré ball with curvature $k$ reads as $P_k^d$ and the proposed fixed-radius Euclidean embeddings as $(S^d, \ell_2)$. The curvatures used in the latter were 0.006, 0.002 in 1s5w, 5s5w (CUB) and 0.002, 0.0002 in 1s5w, 5s5w (Mini), respectively.

As posited, when $d$ increases the Poincaré embeddings approach the effective radius. In fact, plugging $\epsilon=0.01$ in (\ref{eq:radius_eff}) we obtain $r_\mathrm{eff}=4.47$ for $k=-0.05$, $r_\mathrm{eff}=9.99$ for $k=-0.01$ and $r_\mathrm{eff}=14.13$ for $k=-0.005$.  In the CUB dataset, the 5s5w training configuration avoids boundary saturation up to $d=512$, however, the results are worse than the fixed-radius Euclidean architecture for the same dimensions. In MiniImageNet 5s5w, we observe embeddings not at the boundary for the largest dimension $d=1024$ considered. That being the case, the test accuracy is higher than that obtained in lower dimensions with embeddings farther from $r_\mathrm{eff}$. Note that in the two datasets considered, we observed the same Euclidean versus hyperbolic ProtoNet few-shot accuracy gap reported in prior work \cite{Guo2022ClippedClassifiers,Khrulkov2019HyperbolicEmbeddings}, especially in the 1s5w setting. However, the empirical advantage of hyperbolic representations no longer holds once normalized Euclidean embeddings together with the $\ell_2$-norm are introduced in the benchmark. In fact, even when Poincaré embeddings yield better results, the difference w.r.t fixed-radius embeddings is minor. Further, this is verified across all the dimensions considered, unlike \eg, graph or word embeddings where, in low dimensions, hyperbolic space offers a sizeable advantage. 

Finally, results on the effect of magnitude clipping in high-dimensions for the CUB dataset are shown in Table \ref{table:results_clip}. These experiments were performed for $d=1024$, $k=-0.05$ and using Riemannian gradient as mentioned in the original paper. For each clipping value $c$, the corresponding effective ball radius can be obtained by plugging $c$ in the exponential map (\ref{eq:poincare_exp_map}) and taking the $\ell_2$-norm. Thus, for $c=2,3$ and $4$ we have $1.877$, $2.618$ and $3.191$, respectively. We see the saturation effect  occurring, especially for larger clipping values, where the test accuracy is coincidentally higher.

\begin{table}[]
\begin{center}
\begin{tabular}{c|c|cccc}
\toprule
$d$  & Space  & Test acc  & $r_\mathrm{min}$ & $r_\mathrm{avg}$ & $r_\mathrm{max}$ \\ 
\hline
\multirow{4}{*}{$2^7$}  & $(\mathbb{R}^d,\ell_2^2)$ & $59.94 \pm 0.23$ & -       & -      & - \\
                        & $P_{-0.05}^d$             & $64.08 \pm 0.23$ & $4.47$  & $4.47$ & $4.47$  \\
                        & $P_{-0.01}^d$             & $56.46 \pm 0.22$ & $9.98$  & $9.99$ & $9.99$  \\
                        & $(S^d,\ell_2)$            & $\mathbf{66.13 \pm 0.23}$ & $12.91$ & $12.91$ & $12.91$ \\ 
\midrule
\multirow{4}{*}{$2^8$}  & $(\mathbb{R}^d,\ell_2^2)$ & $56.27 \pm 0.23$ & -       & -      & -      \\
                        & $P_{-0.05}^d$             & $65.70 \pm 0.23$ & $4.47$  & $4.47$ & $4.47$ \\
                        & $P_{-0.01}^d$             & $59.74 \pm 0.22$ & $9.99$  & $9.99$ & $9.99$  \\
                        & $(S^d,\ell_2)$            & $\mathbf{66.47 \pm 0.23}$ & $12.91$ & $12.91$ & $12.91$ \\ 
\midrule
\multirow{4}{*}{$2^9$}  & $(\mathbb{R}^d,\ell_2^2)$ & $56.71 \pm 0.23$ & -       & -      & -     \\
                        & $P_{-0.05}^d$             & $66.65 \pm 0.23$ & $4.47$  & $4.47$ & $4.47$ \\
                        & $P_{-0.01}^d$             & $61.52 \pm 0.23$ & $9.99$  & $9.99$ & $9.99$  \\
                        & $(S^d,\ell_2)$            & $\mathbf{67.90 \pm 0.23}$ & $12.91$ & $12.91$ & $12.91$ \\  
\midrule
\multirow{4}{*}{$2^{10}$} & $(\mathbb{R}^d,\ell_2^2)$ & $55.64 \pm 0.23$ & -       & -       & -    \\
                          & $P_{-0.05}^d$             & $67.21 \pm 0.23$ & $4.47$  & $4.47$  & $4.47$  \\
                          & $P_{-0.01}^d$             & $63.85 \pm 0.23$ & $9.99$  & $9.99$  & $9.99$ \\
                          & $(S^d,\ell_2)$            & $\mathbf{67.92 \pm 0.23}$ & $12.91$ & $12.91$ & $12.91$ \\ 
\bottomrule
\end{tabular}
\end{center}
\caption{\textbf{CUB\_200\_2011 1-shot 5-way} test results, 95\% confidence intervals and embedding radii.} 
\label{table:results_cub_1s5w}
\end{table}

\begin{table}[]
\begin{center}
\begin{tabular}{c|c|cccc}
\toprule
$d$  & Space  & Test acc  & $r_\mathrm{min}$ & $r_\mathrm{avg}$ & $r_\mathrm{max}$ \\ 
\hline
\multirow{4}{*}{$2^7$}  & $(\mathbb{R}^d,\ell_2^2)$ & $78.95 \pm 0.16$ & -       & -        & -    \\
                        & $P_{-0.05}^d$             & $79.17 \pm 0.17$ & $4.35$  & $4.47$   & $4.47$   \\
                        & $P_{-0.01}^d$             & $82.30 \pm 0.15$ & $8.07$  & $9.59$   & $9.87$   \\
                        & $(S^d,\ell_2)$            & $\mathbf{83.13 \pm 0.14}$ & $22.36$ &  $22.36$ &    $22.36$       \\ 
\midrule
\multirow{4}{*}{$2^8$}  & $(\mathbb{R}^d,\ell_2^2)$ & $80.14 \pm 0.16$ & -       & -       & -    \\
                        & $P_{-0.05}^d$             & $80.58 \pm 0.16$ & $4.33$  & $4.47$  & $4.47$ \\
                        & $P_{-0.01}^d$             & $84.51 \pm 0.14$ & $9.89$  & $9.99$  & $9.99$   \\
                        & $(S^d,\ell_2)$            & $\mathbf{85.01 \pm 0.14}$ & $22.36$ & $22.36$ &  $22.36$      \\ 
\midrule
\multirow{4}{*}{$2^9$}  & $(\mathbb{R}^d,\ell_2^2)$ & $79.95 \pm 0.15$ & -       & -       & -       \\
                        & $P_{-0.05}^d$             & $81.04 \pm 0.16$ & $4.46$  & $4.47$  & $4.47$  \\
                        & $P_{-0.01}^d$             & $84.60 \pm 0.14$ & $9.90$  & $9.99$  & $9.99$   \\
                        & $(S^d,\ell_2)$            & $\mathbf{85.18 \pm 0.14}$ & $22.36$ & $22.36$ &  $22.36$      \\ 
\midrule
\multirow{4}{*}{$2^{10}$} & $(\mathbb{R}^d,\ell_2^2)$ & $78.83 \pm 0.15$ & -        & -       & -   \\
                          & $P_{-0.05}^d$             & $81.06 \pm 0.16$ & $4.47$   & $4.47$  & $4.47$  \\
                          & $P_{-0.01}^d$             & $84.70 \pm 0.14$ & $9.99$   & $9.99$  & $9.99$   \\
                          & $(S^d,\ell_2)$            & $\mathbf{85.37 \pm 0.14}$ & $22.36$  & $22.36$ & $22.36$  \\ 
\bottomrule
\end{tabular}
\end{center}
\caption{\textbf{CUB\_200\_2011 5-shot 5-way} test results, 95\% confidence intervals and embedding radii.}
\label{table:results_cub_5s5w}
\end{table}

\begin{table}[]
\begin{center}
\begin{tabular}{c|c|cccc}
\toprule
$d$  & Space  & Test acc  & $r_\mathrm{min}$ & $r_\mathrm{avg}$ & $r_\mathrm{max}$ \\ 
\hline
\multirow{4}{*}{$2^7$}  & $(\mathbb{R}^d,\ell_2^2)$ & $49.11 \pm 0.20$ & -       & -      & - \\
                        & $P_{-0.005}^d$            & $49.10 \pm 0.20$ & $3.78$ & $5.85$ & $7.41$  \\
                        & $P_{-0.01}^d$             & $49.60 \pm 0.20$ & $2.78$ & $3.53$ & $4.15$  \\
                        & $(S^d,\ell_2)$            & $\mathbf{50.24 \pm 0.20}$ & $22.36$ & $22.36$ & $22.36$\\ 
\midrule
\multirow{4}{*}{$2^8$}  & $(\mathbb{R}^d,\ell_2^2)$ & $49.14 \pm 0.20$ & -       & -      & -      \\
                        & $P_{-0.005}^d$            & $49.25 \pm 0.20$ & $8.77$ & $9.93$ & $10.44$ \\
                        & $P_{-0.01}^d$             & $47.07 \pm 0.19$ & $7.54$ & $8.05$ & $9.43$  \\
                        & $(S^d,\ell_2)$            & $\mathbf{50.36 \pm 0.20}$ & $22.36$ & $22.36$ & $22.36$\\ 
\midrule
\multirow{4}{*}{$2^9$}  & $(\mathbb{R}^d,\ell_2^2)$ & $48.84 \pm 0.20$ & -       & -      & -     \\
                        & $P_{-0.005}^d$            & $45.59 \pm 0.18$ & $14.12$ & $14.13$ & $14.13$ \\
                        & $P_{-0.01}^d$             & $48.71 \pm 0.19$ & $9.98$  & $9.99$  & $9.99$ \\
                        & $(S^d,\ell_2)$            & $\mathbf{50.97 \pm 0.19}$ & $22.36$ & $22.36$ & $22.36$\\ 
\midrule
\multirow{4}{*}{$2^{10}$} & $(\mathbb{R}^d,\ell_2^2)$ & $49.10 \pm 0.20$ & -       & -       & -    \\
                          & $P_{-0.005}^d$            & $49.19 \pm 0.19$ & $14.13$ & $14.13$ & $14.13$  \\
                          & $P_{-0.01}^d$             & $\mathbf{51.37 \pm 0.20}$ & $9.99$  & $9.99$  & $9.99$ \\
                          & $(S^d,\ell_2)$            & $51.36 \pm 0.20$ & $22.36$ & $22.36$ & $22.36$\\ 
\bottomrule
\end{tabular}
\end{center}
\caption{\textbf{MiniImageNet 1-shot 5-way} test results, 95\% confidence intervals and embedding radii.} 
\label{table:results_mini_1s5w}
\end{table}

\begin{table}[]
\begin{center}
\begin{tabular}{c|c|cccc}
\toprule
$d$  & Space  & Test acc  & $r_\mathrm{min}$ & $r_\mathrm{avg}$ & $r_\mathrm{max}$ \\ 
\hline
\multirow{3}{*}{$2^7$}  & $(\mathbb{R}^d,\ell_2^2)$ & $67.32 \pm 0.16$ & -       & -       & - \\
                        & $P_{-0.005}^d$            & $67.87 \pm 0.16$ & $12.40$ & $13.42$ & $13.69$ \\
                        & $(S^d,\ell_2)$            & $\mathbf{68.57 \pm 0.16}$ & $70.71$  & $70.71$ &  $70.71$\\ 
\midrule
\multirow{3}{*}{$2^8$}  & $(\mathbb{R}^d,\ell_2^2)$ & $67.32 \pm 0.16$ & -       & -      & -      \\
                        & $P_{-0.005}^d$            & $\mathbf{69.40 \pm 0.16}$ &  $13.03$ & $13.78$ & $13.93$ \\
                        & $(S^d,\ell_2)$            & $69.10 \pm 0.16$ & $70.71$ & $70.71$ & $70.71$ \\ 
\midrule
\multirow{3}{*}{$2^9$}  & $(\mathbb{R}^d,\ell_2^2)$ & $66.83 \pm 0.16$ & -       & -       & -     \\
                        & $P_{-0.005}^d$            & $\mathbf{70.20 \pm 0.16}$ & $14.06$ & $14.13$ & $14.13$  \\
                        & $(S^d,\ell_2)$            & $69.95 \pm 0.16$ & $70.71$ & $70.71$ & $70.71$ \\  
\midrule
\multirow{3}{*}{$2^{10}$} & $(\mathbb{R}^d,\ell_2^2)$ & $67.92 \pm 0.16$ & -       & -       & -    \\
                          & $P_{-0.005}^d$            & $\mathbf{70.76 \pm 0.16}$ & $14.06$ & $14.13$ & $14.13$ \\
                          & $(S^d,\ell_2)$            & $70.48 \pm 0.16$ & $70.71$ & $70.71$ & $70.71$ \\ 
\bottomrule
\end{tabular}
\end{center}
\caption{\textbf{MiniImageNet 5-shot 5-way} test results, 95\% confidence intervals and embedding radii.} 
\label{table:results_mini_5s5w}
\end{table}

\begin{table}[]
\begin{center}
\begin{tabular}{c|cccc}
\toprule
Clipping value  & Test acc  & $r_\mathrm{min}$ & $r_\mathrm{avg}$ & $r_\mathrm{max}$ \\ 
\hline
$2.0$ & $62.21 \pm 0.24$ & $1.873$ & $1.877$ & $1.877$\\
$3.0$ & $65.04 \pm 0.24$ & $2.618$ & $2.618$ & $2.618$ \\
$4.0$ & $66.26 \pm 0.23$ & $3.191$ & $3.191$ & $3.191$\\ 
\bottomrule
\end{tabular}
\end{center}
\caption{Effect of feature clipping before the exponential map. \textbf{CUB\_200\_2011 1-shot 5-way} test results, 95\% confidence intervals and embedding radii in $P_k^d$ for $d=1024$ and $k=-0.05$.} 
\label{table:results_clip}
\end{table}

\section{Discussion}
As evidence by our results, in few-shot recognition, Euclidean embeddings do not perform worse than hyperbolic ones, as previously assumed. In fact, they can fare even better. This disparity with respect to the state-of-the-art can be attributed to the fact that the benchmarks conducted in previous papers considered Euclidean networks endowed with the squared $\ell_2$-norm as a distance function, as originally proposed by Snell \etal \cite{Snell2017PrototypicalLearning}. We, on the other hand, used the $\ell_2$-norm and normalized the embeddings' magnitude, resulting in a chordal metric on a sphere.

In order to interpret these results in light of the success that hyperbolic embeddings have had in other areas, one must take into account that the dimensions required to have reasonable few-shot classification accuracies, (typically $d > 10^3$ \cite{Guo2022ClippedClassifiers,Khrulkov2019HyperbolicEmbeddings}), are much higher than those used to embed words \cite{Chamberlain2017NeuralSpace,Nickel2017PoincareRepresentations}, taxonomies or graphs \cite{Chami2019HyperbolicNetworks,Chami2020FromClustering} in general. In addition, while in natural language one may argue that hyperbolicity arises naturally from the co-occurrence of symbols and hence the graph-like nature of the data, this property does not translate directly to the images. For instance, ImageNet is endowed with a taxonomy, inherited from the hypernymy and hyponymy relations in WordNet. The latter has been successfully embedded in low-dimensional hyperbolic space given its symbolic nature \cite{Nickel2018LearningGeometry,Nickel2017PoincareRepresentations,Liu2020HyperbolicRecognition}. However, there may not be a unique visual representation of the abstract concepts in high-levels of this hierarchy. Only labels in leaf-nodes can be ascribed a unique image. In the case of balanced hierarchies or trees, from a symmetry standpoint, the embeddings of these leaf nodes are expected to lie at the same distance from the origin, akin to what is obtained for the image embeddings.

\section{Conclusion}
In this paper we set out to demonstrate that, contrary to widespread belief, hyperbolic embeddings are not necessarily better than Euclidean ones in few-shot learning. We showed that, in high-dimensional Poincaré balls, embeddings tend to saturate at the effective boundary. Consequently, the hierarchy-encoding property of hyperbolic space is lost and, in this situation, whatever accuracy hyperbolic few-shot classifiers have, can be ascribed to the angular separation of their representations. Based on this, we presented empirical results demonstrating that hyperbolic performance on two popular few-shot learning tasks can actually be matched and even improved upon via traditional Euclidean encoders with fixed-radius embeddings. We believe this work can open up new research directions on the design of more effective benchmarks for hyperbolic embeddings, as well as on the identification of computer vision applications wherein hyperbolic geometry presents an unambiguous advantage over Euclidean space.

\paragraph{Acknowledgments} The authors would like to thank the reviewers for their comments and suggestions. This work was funded by Fundação para a Ciência e Tecnologia (FCT), within the scope of the CMU-Portugal Program [iFetch project - LISBOA-01-0247-FEDER-045920] and UIDB/50009/2020. G Moreira was supported by FCT grant contract BL216/2023\_IST-ID and M Marques was supported by the SmartRetail project [PRR - C645440011-00000062], through IAPMEI - Agência para a Competitividade e Inovaç\~ao.

{\small
\bibliographystyle{ieee_fullname}
\bibliography{references}
}


\end{document}